\documentclass{article}

\usepackage[english]{babel}

\usepackage[letterpaper,top=2cm,bottom=2cm,left=3cm,right=3cm,marginparwidth=1.75cm]{geometry}

\usepackage{amsmath}
\usepackage{graphicx}
\usepackage{natbib}
\usepackage[colorlinks=true, allcolors=blue]{hyperref}
\usepackage{graphicx}
\usepackage{url}
\usepackage{hyperref}

\title{Why transformers are \emph{obviously} good models of language}
\author{Felix Hill}

\begin{document}
\maketitle

\begin{abstract}
Nobody knows how language works, but many theories abound. Transformers are a class of neural networks that process language automatically with more success than alternatives, both those based on neural computations and those that rely on other (e.g. more symbolic) mechanisms. Here, I highlight direct connections between the transformer architecture and certain theoretical perspectives on language. The empirical success of transformers relative to alternative models provides circumstantial evidence that the linguistic approaches that transformers embody should be, at least, evaluated with greater scrutiny by the linguistics community and, at best, considered to be the \emph{currently best available} theories. 
\end{abstract}

\section{Word embeddings and lexical prototypes}

\begin{figure}[h]
\label{prototypes}
\centering
\includegraphics[width=15cm, height=7cm]{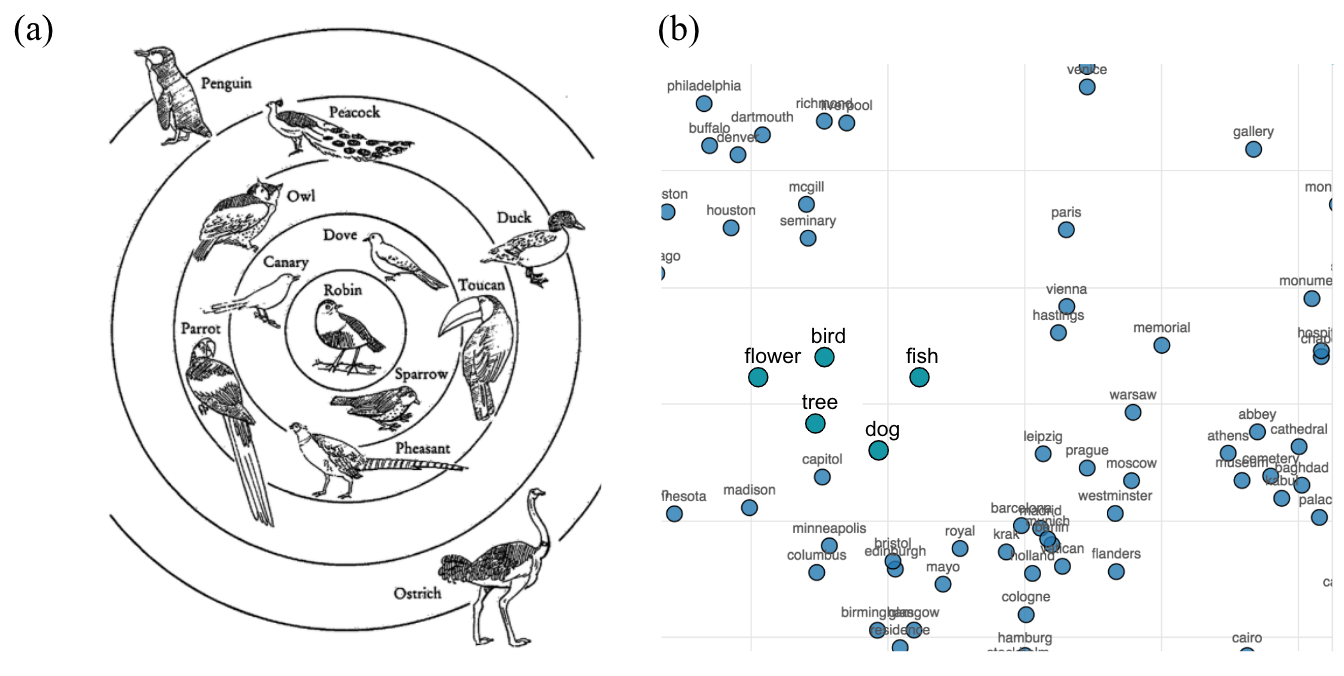}
\caption{(a) An example of prototype structure for the concept \emph{bird} taken from \protect{\citet{rosch1978principles}} A robin is considered by most survey respondents to be a more prototypical bird than is a penguin or ostrich. (b) A 2D (t-SNE) mapping of high dimensional word embeddings learned by the word2vec algorithm~\protect{\citep{mikolov2013efficient}} trained on a large text corpus.
Word embeddings can be uncovered in the input weights of any trained neural language model. Here, I propose we consider them as proxies for prototypical meaning. The range of  different bird variants depicted in (a) would all be centred conceptually on the word-concept \emph{bird} in (b). \footnotesize{[(b) adapted from \url{https://nlpforhackers.io/word-embeddings}]}}

\end{figure}

Among the many unknown things about language is why we give specific names to particular objects or categories in the world around us. \citet{Brown1958HowSA} discussed this question at length, noting that adult speakers of a language typically converge on the names that they give to common objects or categories. Building on this work, \citet{rosch1978principles} proposed that many semantic domains have clear \emph{basic level categories}, such as \emph{apple, fish or knife} (See Fig~\ref{prototypes} (a) for an example). Basic level categories (and their corresponding names) can be more easily learned by infants than their subordinate (\emph{Granny Smith, salmon, cleaver}) or superordinate (\emph{food, animal, tool}) categories. Rosch's work led to the popularisation of the idea that words or concepts can have a \emph{prototypical meaning}. A given instance of a word might sit closer or further from that prototype, depending on the context, and (analogously) different members of a category might sit closer or further from the category's prototypical instance. 

In aiming to connect transformers to theories of human language processing, I propose that, inside a trained transformer, a word (or word piece)'s `embedding' (input) weights should be thought of as reflecting that word (or category)'s prototypical meaning. These weights are activated if and only if a model observes a specific word, word-piece or category name. 

While it may or may not be a new idea to connect word(piece) embeddings in neural networks to Rosch's notion of prototype, the idea of representing a word's most typical meaning according to the average of the contexts in which it appears across a  text corpus dates at least to \cite{cordier-1965-factor}. Later, \cite{miikkulainen1991natural} showed how such a `distributed lexicon' could emerge naturally in a neural language model. As computer hardware developed, \cite{collobert2008unified} proposed that, with a large enough span of words (`vocabulary') these sort of distributed lexical representations could be used to build \emph{unified architectures} for general purpose language processing. See Fig~\ref{prototypes} (b) for an illustration)

\section{Contextualised word meanings}

Unfortunately for those looking for easy solutions, words almost never occur in isolation linguistic or otherwise. In a transformer, a better proxy for a word’s \emph{contextualised} (or `constructed') meaning would be the activations computed after (at least one) of the model's self-attention layers. After the first transformer layer, these activations should capture the `merger' of a word's typical meaning with the prototypical meanings of surrounding words. In other words, a transformer should by default continually update a word's prototypical meaning in order to better estimate its (presumably more appropriate) contextual meaning.

\begin{figure}[h]
\label{embs}
\centering
\includegraphics[width=15cm, height=7cm]{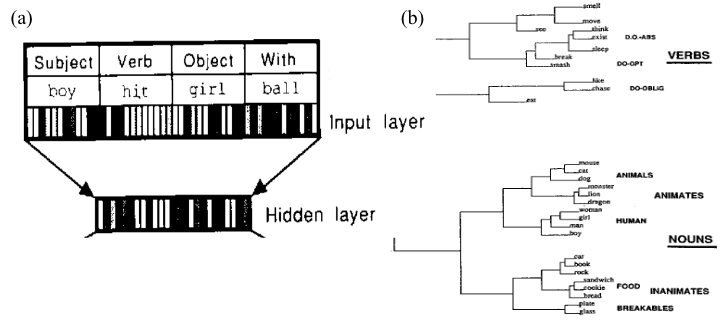}
\caption{\textbf{Early attempts at building on contextualiseed word representations} (a) \protect{\cite{miikkulainen1991natural} } envisaged a `hidden layer' where the meaning of words in different semantic and syntactic contexts would be pooled. (b) Elman fed his simple recurrent network sentences and clustered the resulting internal state at the point immediately following words of interest. The result was semantic clusters emerging naturally from the syntactic patterns build into his synthetic word-like input sequences.} 
\end{figure}

To my knowledge, \cite{elman1990finding} was the first to analyse the semantics of contextualised word embeddings in neural networks. Elman trained small recurrent networks on synthetic symbolic data designed to reflect specific semantic classes, and found that, after some epochs of training, the internal states of these networks clustered naturally into the designated semantic classes. I.e. the network was able to infer semantic information like category membership from syntactic patterns in symbolic text-like input, and to represent this semantic information in internal activations (Fig~\ref{embs} (b)). 

Methods that employ contextualized word embeddings for downstream NLP tasks recruit more of a pretrained network's weights than methods that rely solely on decontextualized word embeddings. In the modern setting of large (web-scale) datasets and extensive pre-training, \cite{peters-etal-2018-deep} demonstrated the value added by doing this. They showed that Elman's method could apply to a LSTM network trained on a massive text corpora - yielding performance across a range of classification-style tasks that was superior to methods based solely on decontextualised word embeddings. 

\section{Contextualised sentence representations}

I hope that the previous section calls into question if it is useful to think of a word having an intrinsic, rather than contextually constructed, meaning (beyond the notion of prototype, at least). This argument, however -- that words do not in themselves carry much meaning -- can be equally made about sentences or any compound linguistic utterance. Like words, sentences are almost always employed in a specific communicative and/or physical context, and almost never employed in isolation. 

The first neural net to capture this idea was the LSTM-based \emph{SkipThought} model \citep{kiros2015skipthought}, which trains LSTMs such that their internal state after processing a sentence is optimal for decoding the subsequent sentence. These `final states' can then be considered as a context-dependent representation of the sentence in question. 

One limitation with SkipThought vectors is that, while they are optimal for capturing contextual meaning, they may be \emph{too} contextual, in that their only purpose is to make decoding extra-sentential content possible. To mitigate this, some colleagues and I worked out enhancements to SkipThought vectors to imbue sentences with both contextual \emph{and} more intrinsic meaning. This involved, for example, using dictionary definitions to align sentence meanings to the meanings that lexicographers understand words to have. We also used images to ensure that sentence representations contained specific concrete information about their relevant visual context \citep{hill-etal-2016-learning}.  

While a partial improvement on SkipThought vectors, these intrinsic-contextual sentence representations were far from perfect. One limitation was the way they were employed; a user must generate a single vector to provide a dowsntream system with appropriate context, and then train (typically linear) layers on top of these vectors in order to make predictions across various NLP tasks.\footnote{At the time, Kyunghyun Cho, Jamie Kiros and I joked about how large the table needed to report these results had to be - something that made the papers themselves hard to process.} The limitation of both SkipThought and our extensions were best expressed by NLP doyen \href{https://www.cs.utexas.edu/~mooney/}{Ray Mooney}, who intuited that it simply shouldn't be possible to \emph{cram the whole meaning of a sentence into a single vector}.  

\begin{figure}[h]
\centering
\label{tuning}
\includegraphics[width=14cm, height=7cm]{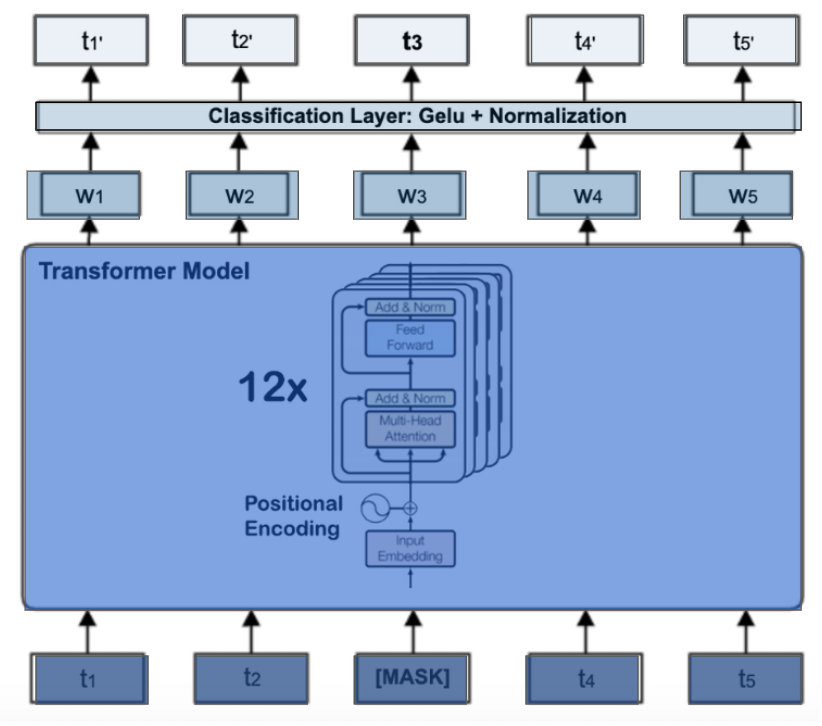}
\caption{\footnotesize{\emph{--adapted from \protect{\cite{khalid2021rubert}}}} -- A simplified illustration of the extent to which different LM-based systems rely on generalized pre-training vs. task-specific tuning. Circa 2008, pre-training only word (prototype) embeddings (depicted in the darkest shade) was common. Following \cite{peters-etal-2018-deep}, the computation of meaning in context (illustrated by the next darkest shade of blue) was exploited directly by downstream systems. Shortly after, applying \protect{\cite{devlin2018bert}}, which required pretraining an entire transformer network, enabled an even wider class of knowledge transfer from generic text corpora to system-specific tasks. Finally, as of 2019, GPT-tyle approaches \protect{\cite{radford2019language}} render pre-training, in some sense at lease, the only game in town. Task-specific behaviour can be achieved in GPT networks with no task-specific weight updates, via a process known as `prompting'.}
\end{figure}

By 2018 BERT \citep{devlin2018bert} had resolved many of these issues. As per \cite{hill-etal-2016-learning}, BERT representations contain both intrinsic meaning, driven by within sentence word prediction, and extrinsic/contextual meaning via a next-sentence prediction objective. Improving on prior work, BERT is also built with a (very large) transformer, and trained on more text than models like SkipThought. Further, BERT makes no commitment about how it should be applied to downstream tasks. There are cases where BERT's representations can be computed a-priori from some context and a linear classifier trained to use those representations. In other cases, however, the whole of BERT can be fine-tuned for optimal predictions. 

One final serendipitous aspect of BERT was that by 2018 it was much easier to demonstrate a model's unequivocal superiority over other models, because instead of building a really large table of results on linear classification tasks and squeezing it into a paper, BERT's authors could just refer to GLUE \citep{wang2018glue}, a more robust and user-friendly web-based multi-task leaderboard. 

See Fig~\ref{tuning} for an illustration of how between 2008 and the present day, NLP systems have come to rely less on low-level intermediate weight vectors and more on flexible distributions of activations from general-purpose networks trained on massive web-scale corpora. Ray Mooney should be rightly impressed by this development. 

The final chapter in this tale is of course the GPT family of models~\citep{radford2019language}. GPT training is both within-sentence and across sentences (thanks to increasingly large transformer memories) so its representations are both intrinsic and contextual. Most crucially, however, GPT models do not need \emph{any} task-specific training once (pre)trained on web-scale corpora. They can just be `prompted' by human users, yielding strong performance on benchmarks like GLUE with no changes to model weights beyond those originally acquired during pretraining.  
 
\section{Top-down effects, constructions and (whisper it) syntax}

In 2015 after working with RNNs for the first time, I made a somewhat brazen statement that \emph{syntax isn't a thing}.\footnote{The correct way to cite this assertion is \emph{Kyunghyun Cho; personal communication}.} This meme ruffled some feathers, but I didn't actually then (and do not actually now) believe that humans have no expectations about how different words should relate to each other. Quite the contrary. What I meant in 2015 was that we should think about new ways of inferring such `expectations’ (which we could refer to as tendencies or patterns of processing) from actual human behavioural data (ie. examples of language use in context) rather than by other methods that were popular at the time. 

In 2015, a popular way to derive a syntactic representation involved introspection by trained linguists (at least to the level of PhD graduates), often operating in small cohorts (those who could be easily recruited by universities). This `expert intuition' approach seemed to me to have two major flaws. First, it was not clear how to avoid bias among cohorts of highly-trained annotators, many of whom seemed to be drawn from populations in a small set of universities worldwide. Second, there seemed to be a vanishing number of cases where systems based on these described intuited representations of syntax actually led to better performance on any deployed language technology. 

I now know (as deep down I did in 2015) that syntax \emph{is} a thing. As language users, we very often expect a sentence to mean something at exactly the time (or very shortly after) we hear that sentence. Later, sometimes, the meaning we arrive at confounds our expectations. These (realised, or confounded) expectations show that human populations do indeed have a strong sense of how words should fit together, and that those expectations are playing a role in our ability to comprehend and use language as efficiently as we do. Indeed, it is these expectations and/or processing tendencies that I think we can (in 2023) usefully refer to as syntax. 

\begin{figure}[h]
\label{trans}
\centering
\includegraphics[width=14cm, height=7cm]{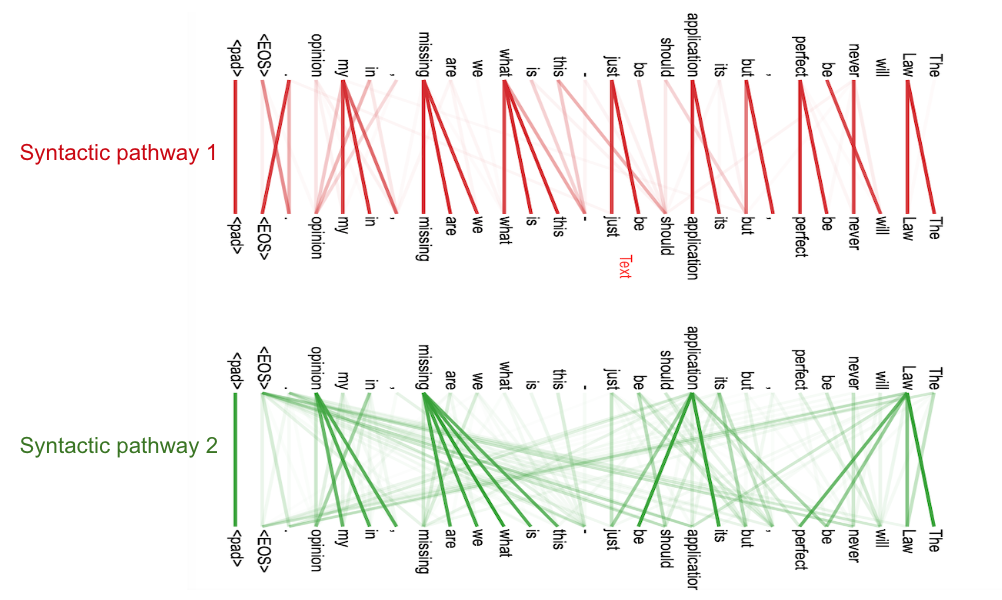}
\caption{As \cite{vaswani2017attention} show, it is easy to plot attention patterns from trained Transformers. However, this alone does not provide us with any nomonological understanding of how transformers achieve such effective construction of phrasal or sentential meaning, distributed across their many network components, on such a diverse set of tasks or contexts.}
\end{figure}

Transformers are great at syntax, because they operate precisely by inferring patterns between input word(pieces) in a way that optimally satisfies their objectives. This fact was not lost on the original developers of the transformer, who included syntactic analyses in the appendix of the seminal \emph{Attention is All You Need} (See Fig~\ref{trans} for examples). Of course, the syntax learned by a transformer will depend on its objective -- the original analyses focused on syntax for machine translation. Understanding the syntax that large scale transformers acquire, across the range of tasks and objectives for which they are trained, in pithy ways that humans can understand, remains a daunting research problem. 

One felicitous correspondence between transformer syntax and human syntax is the propensity for `garden path' behaviour. As humans, sometimes we can only really make sense of a sentence once we have had some time to think and process it, and perhaps even after re-working it either out-loud or in our minds. Here are some examples: 

\begin{itemize}
    \item  \emph{Fruit flies like a banana, time flies like an arrow} 
       \item  \emph{The haystack was necessary because the cloth ripped}  (Example from \citet{Bransford1974ASO} - which makes sense only in the context of skydiving)
\item \emph{They sneezed the foam off the cappuccino} (a construction due to Adele Goldberg, recently studied in the context of BERT \citep{tayyar-madabushi-etal-2020-cxgbert})
\item \emph{Foot Faces Arms Body Head} (Apocryphal UK newspaper headline built entirely from body parts -- requires knowledge of 1980s politics)
\end{itemize}

Transformers are great for this sort of processing problem because they contain multiple parallel attention heads. That means that they can compute a (finite) number of analyses of each input sentence in their activations, prior to ultimately seeing which one is most useful for satisfying the ultimate training (or test) objective. Humans, by contrast, often need explanations or additional information in order to make sense of these sorts of cases, as per the annotations I have added in parentheses.

While humans can and do solve such processing conundrums through external expression and consultation, they can also avoid this by exploiting what are sometimes called \emph{top down effects} of processing. A classic example of top-down processing for language is `coercion' \citep{goldberg2009constructions} - efficiently fitting new (low-level) input words into pre-minted (higher-level) patterns or structures in order to optimally make sense of the input as a whole. 

At first blush, top-down effects can appear to require processing that feeds back down a language-processing architecture, whereas transformers are strictly \emph{feed forward} - they process information only from bottom (words) to top (output predictions). Nonetheless, a transformer of sufficient depth can learn to execute top-down-style processing via a sort of `recurrence in depth’. The attention weights up to layer $N$ of a transformer can be used to compute $h$ hypotheses for the analysis of the sentence. These can then be directly compared to the unadulterated input (via skip connections) to see which in retrospect fits best. This exercise in constraint satisfaction seems to me (squinting somewhat) to be quite analogous to the process of coercion that is so central to Construction-based Linguistics~\citep{goldberg2009constructions} 

Indeed, across a large corpus, I see no reason why the most dominant firing patterns that coalesce in this way (in the first N layers, say, of a transformer) can be considered entirely akin to constructions - the most common patterns of lexical interactions that we as language users come to expect and work with when fitting new words into meaningful utterances.

\section{Many local dependencies, but no local biases?}

One intuitive reason why it is surprising that transformers perform better than recurrent networks on many language tasks is that we know that language is replete with \emph{local dependencies}. It is far more likely that two words are semantically related if they occur in close proximity in some corpus. LSTMs and RNNs contain an inherent bias to discover such connections, by the limitations of back-propagating through time~\citep{pascanu2013difficulty}. 

Transformers, on the other hand, contain no such bias. A transformer will find dependencies between any two words (within its context window of 4000 words, say) with equal prior probability.\footnote{This assumes, of course, that no funky tricks like sinusoidal position encodings are used - but strong transformer performance has been achieved in many cases without these.}

If language is replete with local dependencies and recurrent nets are well suited for learning about such dependencies, why do transformers seem to perform so much better in recent language technology systems? 

First, it's worth noting that the local dependencies noted above are so frequently represented in the data that perhaps `any' model can learn them given enough text. It is plainly evident that transformers can learn these sorts of dependencies. Any advantaged conferred to RNNs vs. Transformers as a consequence of their local dependency bias may be simply too small (or not sufficiently valued by practitioners, in the case of data efficiency) to register at the scale we currently tend to model and evaluate language systems.

Second, by definition if a model has a bias towards local dependencies it has a bias \emph{against} long distance dependencies. Transformers have no bias towards local dependencies and ergo no bias against longer dependencies (within their context window). It is a statistical fact about natural language that long distance dependencies are rarer in the data than short distance dependencies, and a known fact about error-based learning that rarer things are \emph{harder} to learn about than more frequent things. 

Perhaps the lesson here is that, at least for domains of sufficient data, we have reached a state of the art where architectural or structural innovations to neural networks should be considered in light of rare phenomena rather than common ones.

\section{Conclusions}

I have argued that transformers are a good model of language, and that this fact is obvious. But this does not mean they are obviously perfect models of language. Substantial challenges remain. Elsewhere, with colleagues, I have set out some of these challenges and potential solutions~\citep{mcclelland2020placing}.

We know that transformer language models assert false facts with the same degree of authority as they do true information. Fixing this may not be easy, and may require models grounded to a far greater extent in a world in which notions of truth or falsehood can be slowly acquired and then expanded upon via robust inference mechanisms. For language-based assistants, we must also solve the `I insulted your mom problem' - if a human user's conversation with a transformer-based chatbot becomes heated or argumentative on a Friday, it's unlikely to remember or treat you any differently on a Monday. I'll leave it to the reader to consider whether or not this myopia is a desirable trait 
of existing LLMs. 

My principal conclusion however -- and the main motivation for writing this note -- is to point out how many elements of transformers are almost uncannily aligned with certain well-known, but not necessarily well-studied, perspectives on language. I refer particularly (but not exclusively) to those rooted in \href{https://en.wikipedia.org/wiki/Cognitive_linguistics}{Cognitive Linguistics} and \href{https://en.wikipedia.org/wiki/Construction_grammar}{Construction Grammar}. If my intuitions or arguments are correct, and if folks really are impressed with what current LLMs are doing linguistically (if not socially), then maybe that should tell us something about language and linguistics. Namely that historically the ideas of Brown, Shanks, Rosch, Rumelhart, Langacker, Lakoff, Goldberg, McClelland et al. were right, and the views of very many influential others were wrong. Given how many of the above are still actively and passionately pursuing research at the interface of AI and language, we could do worse than heed their advice when thinking about the next steps for both of these rapidly developing fields. 

\section*{Acknowledgements}

Discussions with Steve Piantadosi, Jay McClelland, Adele Goldberg, Stephanie Chan, Murray Shanahan and Ev Fedorenko made this commentary better. Once I share it more widely, others are sure to follow. Please cite it as \citet{hill2023obviously}. 

\bibliographystyle{apalike}
\bibliography{sample}

\end{document}